\documentclass[10pt,twocolumn,letterpaper]{article}

\usepackage{cvpr}
\usepackage{times}
\usepackage{epsfig}
\usepackage{graphicx}
\usepackage{amsmath}
\usepackage{amssymb}
\usepackage{subfigure}

\usepackage[pagebackref=true,breaklinks=true,letterpaper=true,colorlinks,bookmarks=false]{hyperref}

 \cvprfinalcopy 

\def\cvprPaperID{9608} 

\ifcvprfinal\fi

\begin{document}

\title{ SiamCAR: Siamese Fully Convolutional Classification and Regression\\ for Visual Tracking}

\author{Dongyan Guo, Jun Wang, Ying Cui$^*$, Zhenhua Wang, Shengyong Chen\\
	\\
	College of Computer Science and Technology\\
	Zhejiang University of Technology, Hangzhou 310023, China\\
	{\tt\small cuiying@zjut.edu.cn}
}

\maketitle

\begin{abstract}
By decomposing the visual tracking task into two subproblems as classification for pixel category and regression for object bounding box at this pixel, we propose a novel fully convolutional Siamese network to solve visual tracking end-to-end in a per-pixel manner. The proposed framework SiamCAR consists of two simple subnetworks: one Siamese subnetwork for feature extraction and one classification-regression subnetwork for bounding box prediction. Our framework takes ResNet-50 as backbone. Different from state-of-the-art trackers like Siamese-RPN, SiamRPN++ and SPM, which are based on region proposal, the proposed framework is both proposal and anchor free. Consequently, we are able to avoid the tricky hyper-parameter tuning of anchors and reduce human intervention. The proposed framework is simple, neat and effective. Extensive experiments and comparisons with state-of-the-art trackers are conducted on many challenging benchmarks like GOT-10K, LaSOT, UAV123 and OTB-50. Without bells and whistles, our SiamCAR achieves the leading performance with a considerable real-time speed. 
\end{abstract}


\section{Introduction}
Visual object tracking has received considerable attention due to its wide application such as intelligent surveillance, human-machine interaction and unmanned vehicles.
Rapid progress has been made on visual tracking. However, it remains a challenging task especially for real world applications, as object in unconstrained recording conditions often suffers from large illumination variation, scale variation, background clutters and heavy occlusions, etc. Moreover, the appearance of non-rigid objects may change significantly due to extreme pose variation.

\begin{figure}
	\centering 
	\includegraphics[scale=0.13]{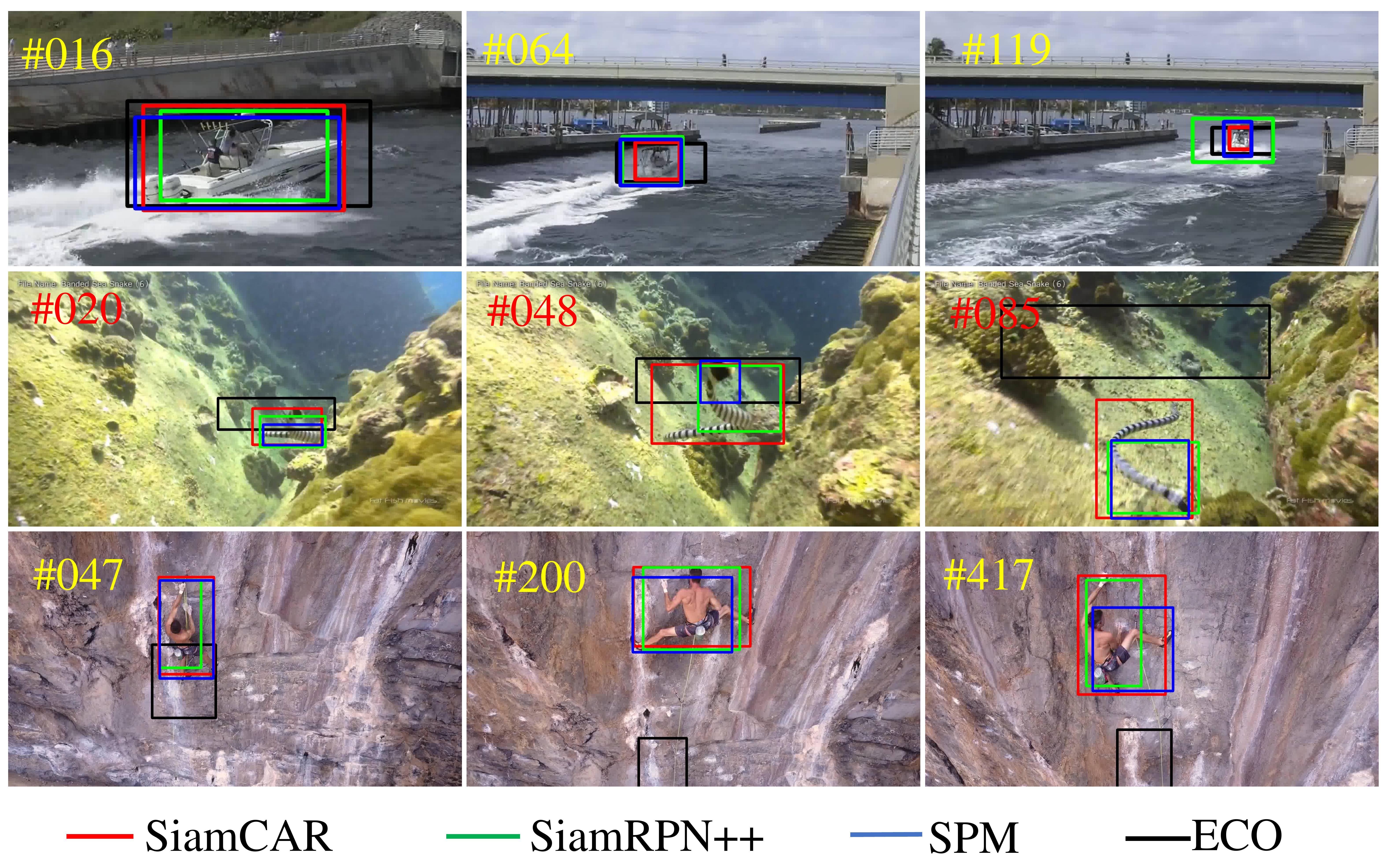}
	\caption{\footnotesize{Comparisons of the proposed SiamCAR with three state-of-the-art trackers on three challenging sequences from GOT-10K. Our SiamCAR can accurately predict the bounding boxes even the objects suffer from similar distractors, large scale variation and large pose variation, while SiamRPN++ and SPM give much rougher results and ECO drifts to the background. }}
	\label{fig1}
\end{figure}

The current popular visual tracking methods \cite{SiamFC, SiamRPN, saSiam,cfnet,siamrpn++,DaSiamRPN,CSiam, song2017crest} revolve around the Siamese network based architectures. The Siamese network formulates the visual tracking task as a target matching problem and aims to learn a general similarity map between the target template and the search region. 
Since one single similarity map typically contains limited spatial information, 
a common strategy is to perform matching on multiple scales of the search regions to determine the object scale variation \cite{SiamFC, saSiam,cfnet}, which explains why these trackers are time-consuming and labor-intensive. 
SiamRPN \cite{SiamRPN} attaches the Siamese network a region proposal extraction subnetwork (RPN). 
By jointly training a classification branch and a regression branch for visual tracking, SiamRPN avoids the time-consuming step of extracting multi-scale feature maps for the object scale invariance. It achieves state-of-the-art results on many benchmarks. Later works such as DaSiam  \cite{DaSiamRPN} , CSiam  \cite{CSiam} and  SiamRPN++  \cite{siamrpn++} improves SiamRPN. However, since anchors are introduced for region proposal, these trackers are sensitive to the numbers, sizes and aspect ratios of anchor boxes, and expertise on hyper-parameter tuning is crucial to obtain successful tracking with these trackers. 

In this paper, we show that an anchor-free Siamese network based tracker can perform better than the state-of-the-art RPN based trackers. Essentially we decompose tracking into two subproblems: one classification problem and one regression task. The classification branch aims to predict each spatial location a label, while the regression branch considers regressing each location a relative bounding box.
With such decomposition, the tracking task can be solved in a per-pixel prediction manner. We then craft a simple yet effective Siamese based classification and regression network (SiamCAR) to learn the classification and regression models simultaneously in an end-to-end manner.

Previous work \cite{atom} leverages object semantic information to improve the bounding box regression. Inspired by this, SiamCAR is designed to extract response maps which include affluent category information and semantic information. Different from RPN models \cite{SiamRPN, DaSiamRPN, siamrpn++}, which use two response maps for region proposal detection and regression respectively, SiamCAR takes one unique response map to predict object location and bounding box directly.

\begin{figure*}
	\centering 
	\includegraphics[scale=0.532]{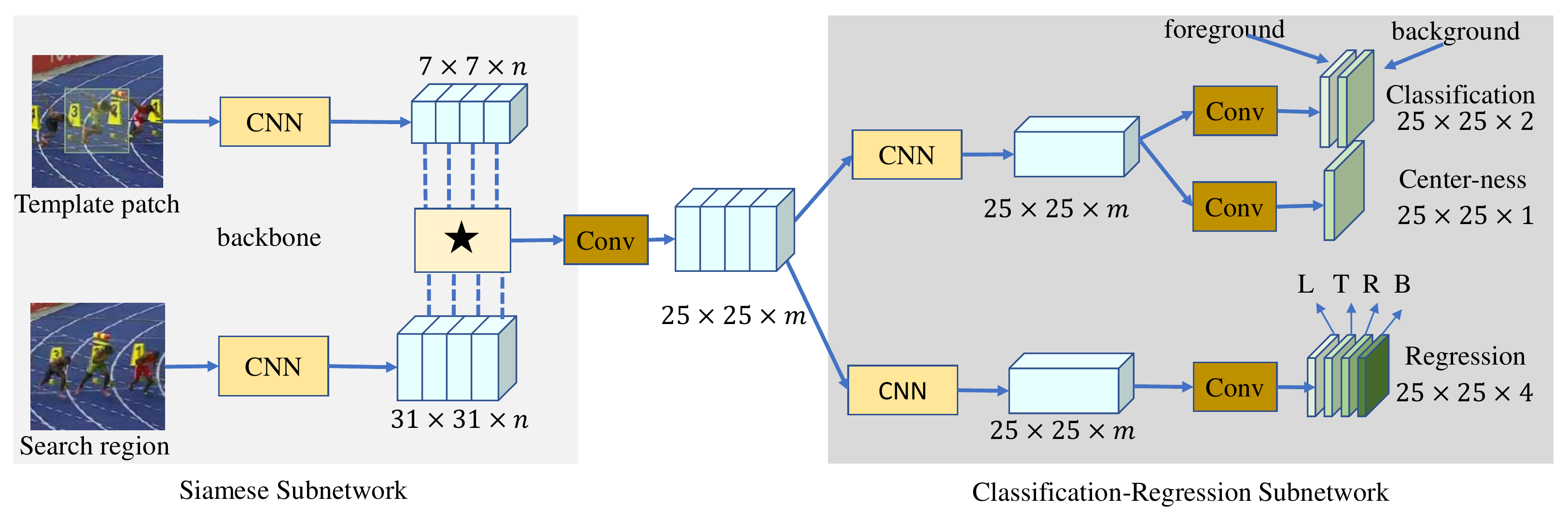}
	
	\caption{\footnotesize{Illustration of SiamCAR: The left side is a Siamese subnetwork with a depth-wise cross correlation layer (denoted by $\bigstar$) for multi-channel response map extraction. The right side shows the classification and regression subnetwork for bounding box prediction, which is taken to decode the location and scale information of the object from the multi-channel response map. Note that SiamCAR can be implemented as a fully convolutional network, which is simple, neat and easy to interpret.}}
	\label{fig-framwork}
\end{figure*}

SiamCAR adopts the strategy of online training and offline tracking, without using any data enhancement during training. Our main contributions are:
\begin{itemize}
\item	We propose the so-called Siamese classification and regression framework (SiamCAR) for visual tracking. The framework is very simple in construction but powerful in performance. 

\item  The proposed tracker is both anchor and proposal free. The number of hyper-parameters has been significantly reduced, which keeps the tracker from complicated parameter tuning and make the tracker significantly simpler, especially in training.

\item  	Without bells and whistles, the proposed tracker achieves the state-of-the-art tracking performance in terms of both accuracy and time cost.

\end{itemize}

\section{Related Works}
We mainly review the family of Siamese RPN trackers since they dominate the tracking performance in recent years.

Tracking researchers devote to design faster and more accurate trackers from different aspects like feature extraction  \cite{henriques2014high,possegger2015defense}, template updating \cite{cfnet, gao2019graph}, classifier design \cite{zhang2017robust} and bounding box regression \cite{atom}.  Early feature extraction mainly uses color features, texture features or other hand-crafted ones.
Benefit by the development of deep learning, now the deep convolutional feature CNN is widely adopted. Template updating can improve the model adaptability, but online tracking is very inefficient. Besides, the tracking drift problem for template updating still need to be solved. The introduction of correlation filter methods \cite{bolme2010visual, danelljan2016discriminative, henriques2014high, li2017integrating, zhang2017multi, liu2016part}  make the tracking reach an unprecedented height both in efficiency and accuracy. The current researches show that the Siamese based online training and offline tracking methods with deep neural network have achieved the best balance between accuracy and efficiency \cite{SiamRPN, siamrpn++}.

As one of the pioneering works, SiamFC \cite{SiamFC} constructs a fully convolutional Siamese network to train a tracker. Encouraged by its success, many researchers follows the work and propose some updated models \cite{dong2018triplet,cfnet, saSiam, DSiam, SiamRPN, siamrpn++} . CFNet \cite{cfnet} introduces the Correlation Filter layer to the SiamFC framework and performs online tracking to improve the accuracy. By modifying the Siamese branches with two online transformations, DSiam \cite{DSiam} proposes to learn a dynamic Siamese network, which achieves an improved accuracy with acceptable speed loss. The SA-siam \cite{saSiam} builds a twofold Siamese network with a semantic branch and an appearance branch. The two branches are trained separately to keep the heterogeneity of features but combined at the testing time to improve the tracking accuracy. In order to deal with the scale variation problem, these Siamese networks need to process multi-scale searching and result in time-consuming problem. 

Inspired by the region proposal network for object detection  \cite{ren2015faster}, the SiamRPN \cite{SiamRPN} tracker  performs the region proposal extraction after the Siamese network outputs. By jointly training a classification branch and a regression branch for region proposal, SiamRPN avoids the time-consuming step of extracting multi-scale feature maps for the object scale invariance and achieves very efficient results. However, it has difficulty in deal with distractors with similar appearance to the object. Based on SiamRPN, DaSiamRPN \cite{DaSiamRPN} increases the hard negative training data during the training phase. Through data enhancement, they improve the discrimination of the tracker and obtain a much more robust result. The tracker is further extended to long-term visual tracking. Up to now the framework has been modified a lot from SiamFC, but the performance still can not move on with deeper network by using AlexNet as backbone. Aims to this problem, SiamRPN++ \cite{siamrpn++} optimizes the network architecture by using the ResNet \cite{resnet} as backbone. At the same time, they randomly shift the training object location in the search region during model training to eliminate the center bias. After these modifications, the better tracking accuracy can be achieved in a very deep network architecture instead of shallow neural networks.

Anchors are adopted in these RPN based trackers for region proposal. Besides, anchor boxes can make use of the deep feature maps and avoid repeated computation, which can significantly speed up the tracking process. The state-of-the-art trackers SPM \cite{spm} and SiamRPN \cite{SiamRPN} both work in a very high speed. Though SiamRPN++ \cite{siamrpn++} adopts a very deep neural network, it can still work in a considerable real-time speed. The accuracy and speed of the state-of-the-art Anchor-free trackers like ECO \cite{ECO} still has gap with these anchor-based trackers \cite{spm, siamrpn++} on the challenging benchmarks like GOT-10K \cite{got10k}. However, the tracking performance is very sensitive to the relative hyper-parameters of anchors, which need to be carefully tuned and empirical tricks are involved to achieve ideal performance. Moreover, since the size and aspect ratio of anchor boxes are fixed, even with heuristic tuned parameters, these trackers still have difficulty in processing objects with large shape deformation and pose variation. In this paper, we show that the problems can be greatly alleviated with our proposed SiamCAR. Moreover, we demonstrate that a tracker with much simpler construction can achieve even better performance than state-of-the-art ones.

\section{Proposed Method}
We now introduce our SiamCAR network in detail. As mentioned, we decompose the tracking task into two subproblems as classification and regression, and then solve them in a per-pixel manner. As show in Figure \ref{fig-framwork}, the framework mainly consists of two simple subnetworks: a Siamese network for feature extraction along with a classification and regression network for bounding box prediction.

\subsection{Feature Extraction with Siamese Subnetwork}
Here we take advantage of the fully convolution network without padding to construct the Siamese subnetwork for the visual feature extraction. The subnetwork consists of two branches: a target branch which takes the tracking template patch $ Z $ as input, and a search branch which takes the search region $ X $ as input. The two branches share the same CNN architecture as their backbone models, which output two feature maps $ \varphi(Z) $ and $ \varphi(X) $. In order to embed the information of these two branches, a response map $ R $ can be obtained by performing the cross-correlation on $ \varphi(X) $ with $ \varphi(Z) $ as a kernel. Since we need to decode the response map $ R $ in the subsequent prediction subnetwork to obtain the location and scale information of the target, we hope that $ R $ retains abundant information.
However, the cross-correlation layer can only generate a single-channel compressed response map, which lacks useful features and important information for tracking, as suggested by \cite{siamrpn++} that different feature channels typically take distinct semantic information. Inspired by  \cite{siamrpn++}, we also use a depth-wise correlation layer to produce multiple semantic similarity maps:
\begin{equation}
R=\varphi(X) \star \varphi(Z),
\end{equation}
where $ \star $ denotes the channel-by-channel correlation operation. The generated response map $ R $ has the same number of channels as $ \varphi(X) $, and it contains massive information for classification and regression.

Low-level features like edge, corner, color and shape that represent better visual attributes are indispensable for location, while high-level features have better representation on semantic attributes and they are more crucial for discrimination. Many methods take advantage of fusing both low-level and high-level features to improve the tracking accuracy \cite{ma2018robust, siamrpn++}. Here we also consider to aggregate multi-layer deep features for tracking. We use the modified ResNet-50 as the same in \cite{siamrpn++} as our backbone networks. To achieve better inference for recognition and discrimination, we compound the features extracted from the last three residual blocks of the backbone, The three outputs denoted respectively as $\mathcal{F}_3(X)$, $\mathcal{F}_4(X)$, $\mathcal{F}_5(X)$ are concatenated as a unity
\begin{equation}
\varphi(X)= Cat(\mathcal{F}_3(X),\mathcal{F}_4(X),\mathcal{F}_5(X)),
\end{equation}
where $\mathcal{F}_{i=3:5}(X)$ includes $256$ channels. Hence  $\varphi(X)$ contains $3\times 256$ channels. 

The Depth-wise Cross Correlation is performed between the searching map $\varphi(X)$ and the template map $\varphi(Z)$ 
to get a multi-channel response map. 
The response map is then convoluted with a $1*1$ kernel to reduce its dimension to $256$ channels. Through the dimension-reduction, the number of parameters can be significantly reduced, in a result the following computation can be speed up. The final dimension-reduced response map $R^{*}$ is adopted as the input to the classification-regression subnetwork.

\subsection{Bounding Box Prediction with Classification and Regression Subnetwork}
Each location $(i,j)$ in the response map $R^{*}$ can be mapped back onto the input search region as $(x,y)$. The {RPN-based} trackers consider the corresponding location on the search region as the center of multi-scale anchor boxes, and regress the target bounding box with these anchor boxes as references. Different from them, our network directly classifies and regresses the target bounding box at each location. The associated training can be accomplished by the fully convolution operation in an end-to-end fashion, which avoids tricky parameter tuning and reduces human intervention.

The tracking task is decomposed into two subtasks: a classification branch to predict the category for each location, and a regression branch to compute the target bounding box at this location (see Figure \ref{fig-framwork} for an illustration of the subnetwork).  For a response map $R^*_{w \times h \times m} $ extracted using the Siamese subnetwork, the classification branch outputs a classification feature map $ A^{cls}_{w \times h \times 2} $ and the regression branch outputs a regression feature map $ A^{reg}_{w \times h \times 4} $. Here $w$ and $h$ represent the width and the height of the extracted feature maps respectively. As that shown in Figure  \ref{fig-framwork}, each point $(i,j,:)$  in  $ A^{cls}_{w \times h \times 2} $ contains a $2D$ vector, which represents the foreground and background scores of the corresponding location in the input search region. Similarly, each point $(i,j,:) $ in $ A^{reg}_{w \times h \times 4} $ contains a $4D$ vector $ t(i,j)=(l,t,r,b)  $, which represents the distances from the corresponding location to the four sides of the bounding box in the input search region.

Since the ratio of areas occupied by the target and the background in the input search region is not very large, sample imbalance is not a problem. Therefore, we simply adopt the cross-entropy loss for classification and the IOU loss for regression. Let $(x_0, y_0) $ and $ (x_1,y_1) $ denote the left-top and right-bottom corner of the ground truth bounding box and $(x,y) $ denote the corresponding location of point $(i,j)$, the regression targets $ \tilde{t}_{(i,j)} $ at $ A^{reg}_{w \times h \times 4}(i,j,:) $ can be calculated by:
\begin{equation}
\begin{split}
&\tilde{t}^{0}_{(i,j)}=\tilde{l}=x-x_0 ,\tilde{t}^{1}_{(i,j)}=\tilde{t}=y-y_0\\
&\tilde{t}^{2}_{(i,j)}=\tilde{r}=x_1-x, \tilde{t}^{3}_{(i,j)}=\tilde{b}=y_1-y\\
\end{split}
\end{equation}
With $ \tilde{t}_{(i,j)} $, the IOU between the ground-truth bounding box and the predicted bounding box can be computed. Then we compute the regression loss by using
\begin{equation}
\mathcal L_{reg}=\frac{1}{\sum{\mathbb{I}}(\tilde{t}_{(i,j)})} \sum\nolimits_{i,j}\mathbb{I}(\tilde{t}_{(i,j)})L_{IOU}(A^{reg}(i,j,:),\tilde{t}_{(x,y)})
\end{equation}
where $ L_{IOU} $ is the IOU loss as in \cite{yu2016unitbox} and $ \mathbb{I}(\cdot ) $ is an indicator function defined by:
\begin{equation}
\mathbb{I}(\tilde{t}_{(i,j)})=\left\{
\begin{array}{lr}
1 & if\  \tilde{t}^{k}_{(i,j)}>0 , k=0,1,2,3\\
0 & otherwise\\
\end{array}\right.
\end{equation}

An observation is that the locations far away from the object center tend to produce low-quality predicted bounding boxes, which reduces the performance of the tracking system. Following \cite{fcos}, we add a center-ness branch in parallel with the classification branch to remove the outliers. As shown in Figure \ref{fig-framwork}, the branch outputs a center-ness feature map $ A^{cen}_{w \times h \times 1} $, where each point value gives the center-ness score of the corresponding location. The score $ C(i,j) $ in $ A^{cen}_{w \times h \times 1}(i,j) $ is defined by
\begin{equation}
C(i,j)=\mathbb{I}(\tilde{t}_{(i,j)}) \ast \sqrt{\frac{min(\tilde{l},\tilde{r})}{max(\tilde{l},\tilde{r})} \times \frac{min(\tilde{t},\tilde{b})}{max(\tilde{t},\tilde{b})}}
\end{equation}
where $ C(i,j) $ is in contrast with the distance between the corresponding location $ (x,y) $ and the object center in the search region. If $ (x,y) $ is located in the background, the value of $ C(i,j) $ is set to 0.
The center-ness loss is
\begin{equation}
\begin{split}
\mathcal L_{cen}= &\frac{-1}{\sum{\mathbb{I}}(\tilde{t}_{(i,j)})} \sum\nolimits_{\mathbb{I}(\tilde{t}_{(i,j)})==1}C(i,j) \ast \log A^{cen}_{w \times h \times 1}(i,j)\\&+(1-C(i,j)) \ast \log(1 - A^{cen}_{w \times h \times 1}(i,j))
\end{split}
\end{equation}

The overall loss function is
\begin{equation}
\mathcal L=\mathcal L_{cls}+ \lambda_1\mathcal L_{cen} +\lambda_2\mathcal L_{reg}
\end{equation}
where $ \mathcal L_{cls} $ represents the cross-entropy loss for classification. Constants $ \lambda_1 $ and $ \lambda_2 $ weight center-ness loss and regression loss. During model training, we empirically set $ \lambda_1 = 1 $ and $ \lambda_2 = 3 $.

\subsection{The Tracking Phase}
Tracking aims at predicting a bounding box for the target in current frame. For a location $ (i,j) $, the proposed framework can produce a 6D vector $ T_{ij}=(cls, cen, l, t, r, b) $, where $ cls $ represents the foreground score of classification, $ cen $ represents the center-ness socre, $ l+r $ and $ t+b $ represent the predicted width and height of the target in current frame. During tracking, the size and aspect ratio of the bounding box typically see minor change across consecutive frames. To supervise the prediction using this spatial-temporal consistency, we adopt a scale change penalty $ p_{ij} $ as that introduced in \cite{SiamRPN} to re-rank the classification score $ cls $, which admits an updated 6D vector $ PT_{ij}=(cls_{ij} \times p_{ij}, cen, l, t, r, b) $. Then the tracking phase can be formulated as:
\begin{equation}
q=\arg\max_{i,j}\{(1-\lambda_d)cls_{ij} \times p_{ij} + \lambda_d{H}\}
\label{getq}
\end{equation}
where $ H $ is the cosine window and $ \lambda_d $ is the balance weight. The output $ q $ is a queried location with the highest score to be a target pixel.

\begin{figure}
	\centering 
	\includegraphics[width=0.5\textwidth]{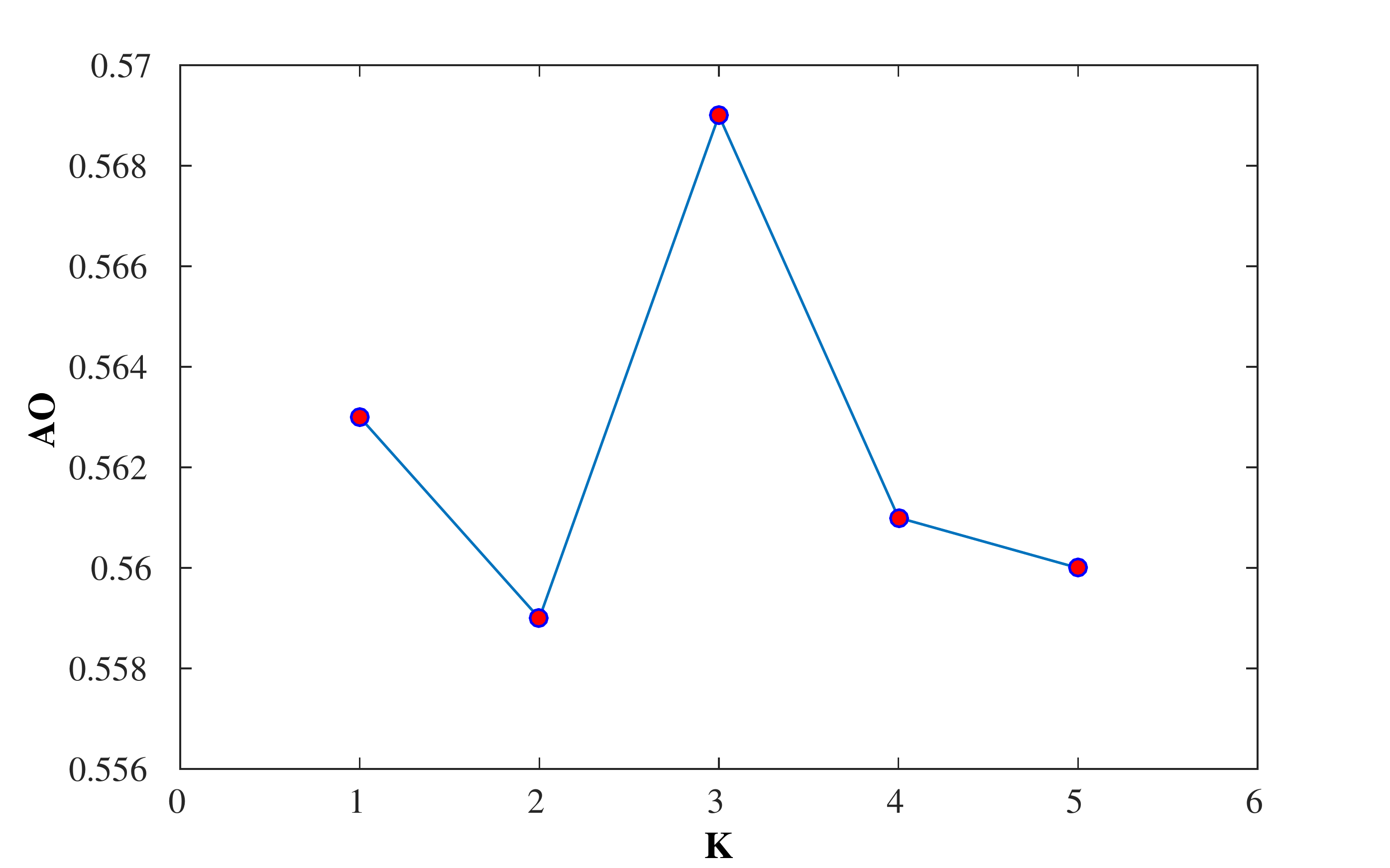}
	\caption{Results on GOT-10K \cite{got10k} using different k values.}
	\label{topk}
\end{figure}

Since our model solves the object tracking with a per-pixel prediction manner, each location is relative to a predicted bounding box. In the real tracking process, it will be jittering between adjacent frames if the only bounding box of $ q $ is used as the target box. We observed that the pixels located around $ q $ are more likely to be the target pixel. Hence we choose the top-k points from $ n $ neighborhoods of $ q $ according to the value $ cls_{ij} \times p_{ij} $. The final prediction is the weighted average of the selected $ k $ regression boxes. Empirically, we found that setting $ n=8 $ and $ k=3 $ delivers stable tracking results (see Figure \ref{topk} for a comparison of using different $k$ values).

\begin{figure}
	\centering 
	\includegraphics[width=0.45\textwidth]{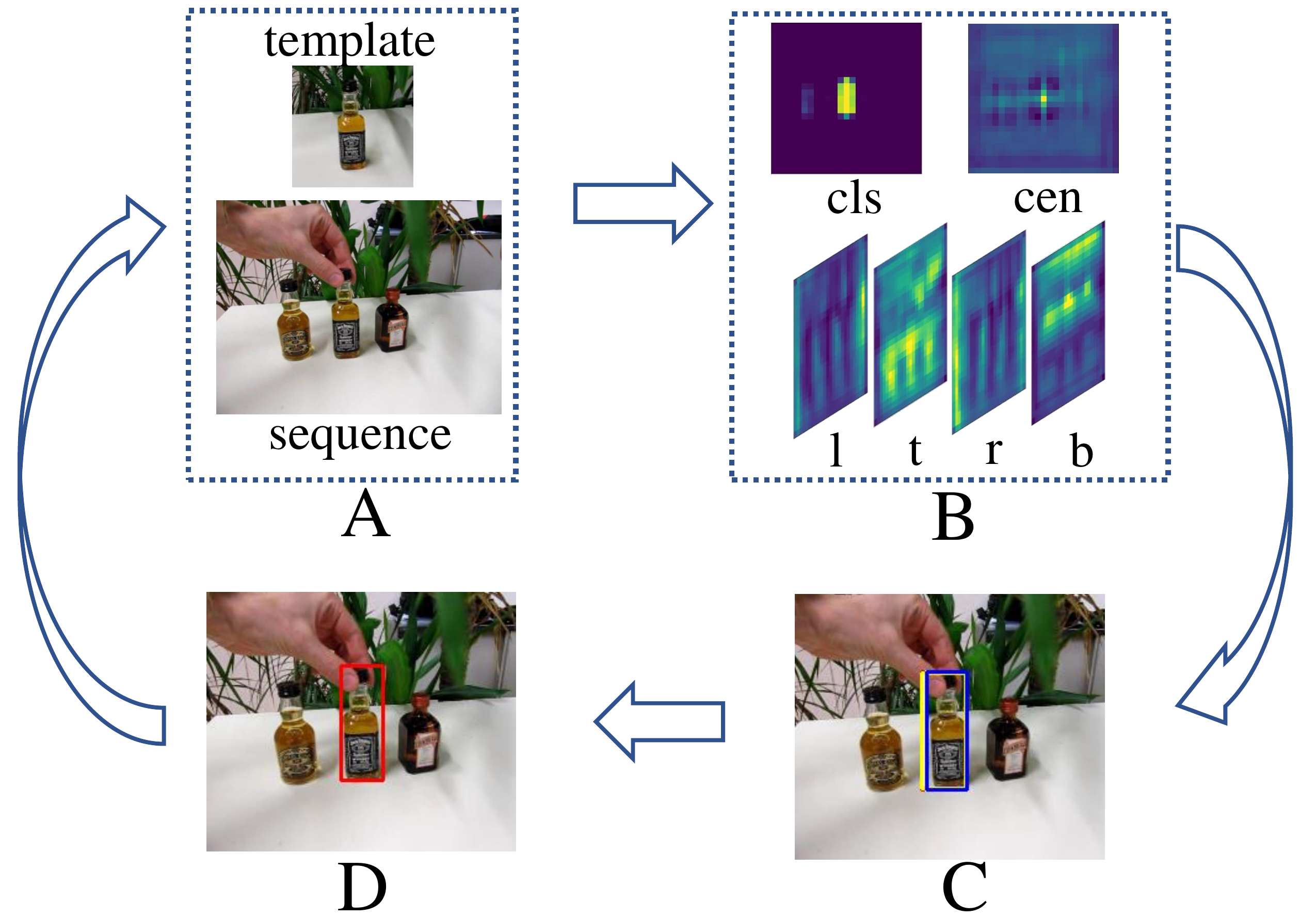}
	\caption{Tracking process: Sub-figure A shows a pair of inputs while Sub-figure B shows the corresponding outputs of the model. From Sub-figure B we can observe that the outputs of the model can depict good prediction for different attributes of the object. Sub-figure C shows the predicted bounding boxes corresponding to the Top-K points. Sub-figure D shows the final predicted bounding box by averaging those boxes in C.}
	\label{trackflow}
\end{figure}

\section{Experiments}
\subsection{Implementation details}
The proposed SiamCAR is implemented in Python with Pytorch on 4 RTX2080ti. For easy comparison, the input size of the template patch and search regions are set as the same with \cite{siamrpn++}, respectively to $127$ pixels and $255$ pixels. The modified ResNet-50 as in \cite{siamrpn++} is adopted as the backbone Siamese subnetwork. The network is pretrained on ImageNet \cite{russakovsky2015imagenet} and then using the parameters as initialization to retrain our model.

\begin{figure}
	\centering 
	\includegraphics[width=0.48\textwidth]{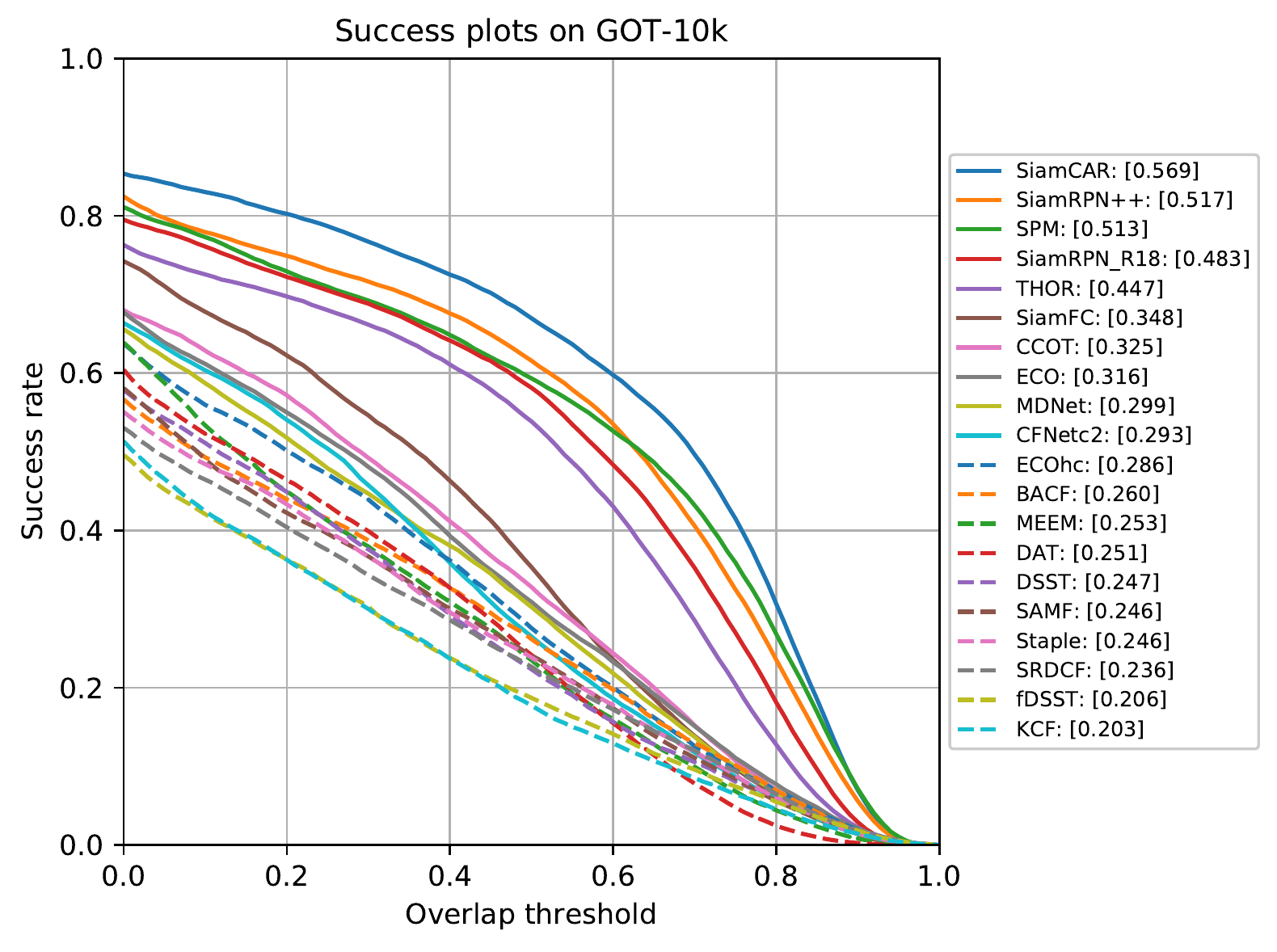}
	\caption{Comparisons on GOT-10K \cite{got10k}. Our SiamCAR significantly outperforms the baselines and other state-of-the-art methods.}
\end{figure}

\begin{table*}
	\begin{center}
		\setlength{\tabcolsep}{5.75mm}
		\begin{tabular}{|l|c|c|c|c|c|c|c|}
			\hline
			Tracker & $AO$ & $SR_{0.5}$ & $SR_{0.75}$& $FPS$ & $Hardware$ &$Language$ \\
			\hline\hline
			KCF \cite{henriques2014high} & 0.203& 0.177& 0.065& 94.66 & CPU & Matlab\\
			fDSST \cite{danelljan2016discriminative}&	0.206&	0.187&	0.075& 30.43 &CPU & Matlab\\
			SRDCF \cite{danelljan2015learning}&	0.236&	0.227&	0.094 &5.58 &CPU &Matlab\\
			Staple \cite{staple}& 0.246& 0.239& 0.089& 28.87& CPU & Matlab\\
			SAMF \cite{SAMF}& 0.246& 0.241 &0.084 & 7.43 &CPU &Matlab\\
			DSST \cite{DSST}& 0.247&0.223 &0.081 &18.25 &CPU& Matlab \\
			DAT \cite{DAT}& 0.251 &0.242 &0.048 &45.52 &CPU &Matlab\\
			MEEM \cite{MEEM}& 0.253 &0.235 & 0.068 &20.59& CPU &Matlab\\
			BACF \cite{BACF}& 0.260 &0.262 &0.101 &14.44 &CPU &Matlab\\
			ECO-HC& 0.286 & 0.276 &0.096 &44.55 & CPU & Matlab\\
			CFnet \cite{cfnet}&0.293	&0.265&	0.087 & 35.62 & Titan X& Matlab\\
			MDnet \cite{mdnet}&	0.299&	0.303&	0.099 &1.52 &Titan X & Python\\
			ECO \cite{ECO}&0.316&0.309&	0.111 &2.62 &CPU &Matlab\\
			CCOT \cite{CCOT}&0.325 & 0.328 &0.107 &0.68 & CPU & Matlab\\
			SiamFC \cite{SiamFC}&0.374&0.404	&0.144 &25.81 & Titan X & Matlab\\
			THOR&	0.447&	0.538&	0.204& 1.00 &RTX 2070& Python\\
			SiamRPN\_R18 &0.483 & 0.581& 0.270& 97.55 & Titan X & Python\\
			SPM \cite{spm}&	0.513&	0.593&	\textbf{\textcolor{blue}{0.359}}& 72.30 & Titan Xp &Python\\
			SiamRPN++ \cite{siamrpn++}&\textbf{\textcolor{blue}{	0.517}}&	\textbf{\textcolor{blue}{0.616}} &	0.325& 49.83 & RTX 2080ti& Python\\
			\hline
			SiamCAR&\textbf{\textcolor{red}{0.569}}	&\textbf{\textcolor{red}{0.670}}	&	\textbf{\textcolor{red}{0.415}}& 52.27 & RTX 2080ti& Python\\
			\hline
		\end{tabular}
	\end{center}
	\caption{Comparisons on GOT-10K \cite{got10k}. The best two results are highlighted respectively in red and blue fonts.}
	\label{compGOT10K}
\end{table*}

\begin{figure*}[htbp]
	\centering
	\subfigure{
		\begin{minipage}[t]{0.31\linewidth}
			\centering
			\includegraphics[width=2.38in]{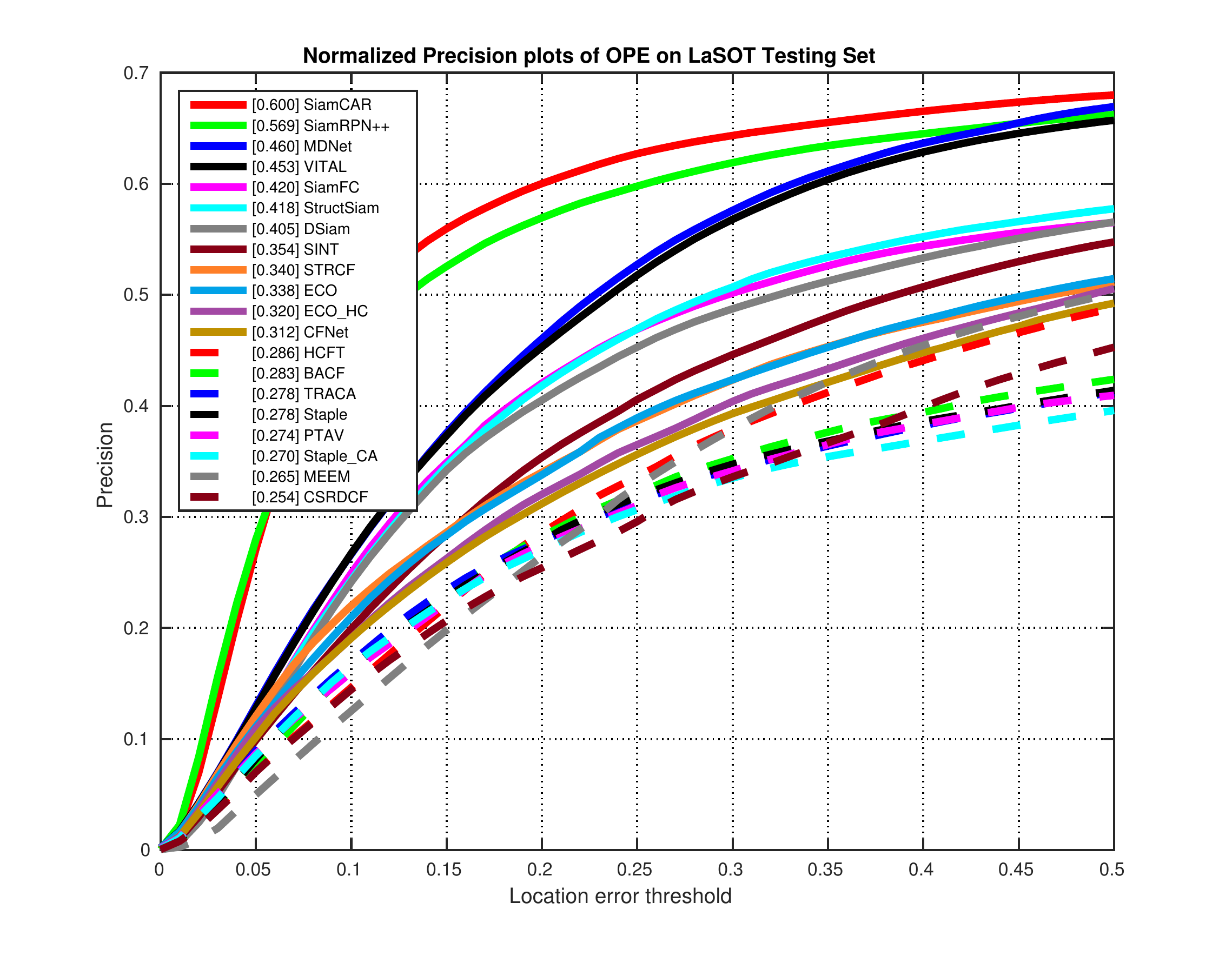}
		\end{minipage}%
	}%
	\subfigure{
		\begin{minipage}[t]{0.31\linewidth}
			\centering
			\includegraphics[width=2.38in]{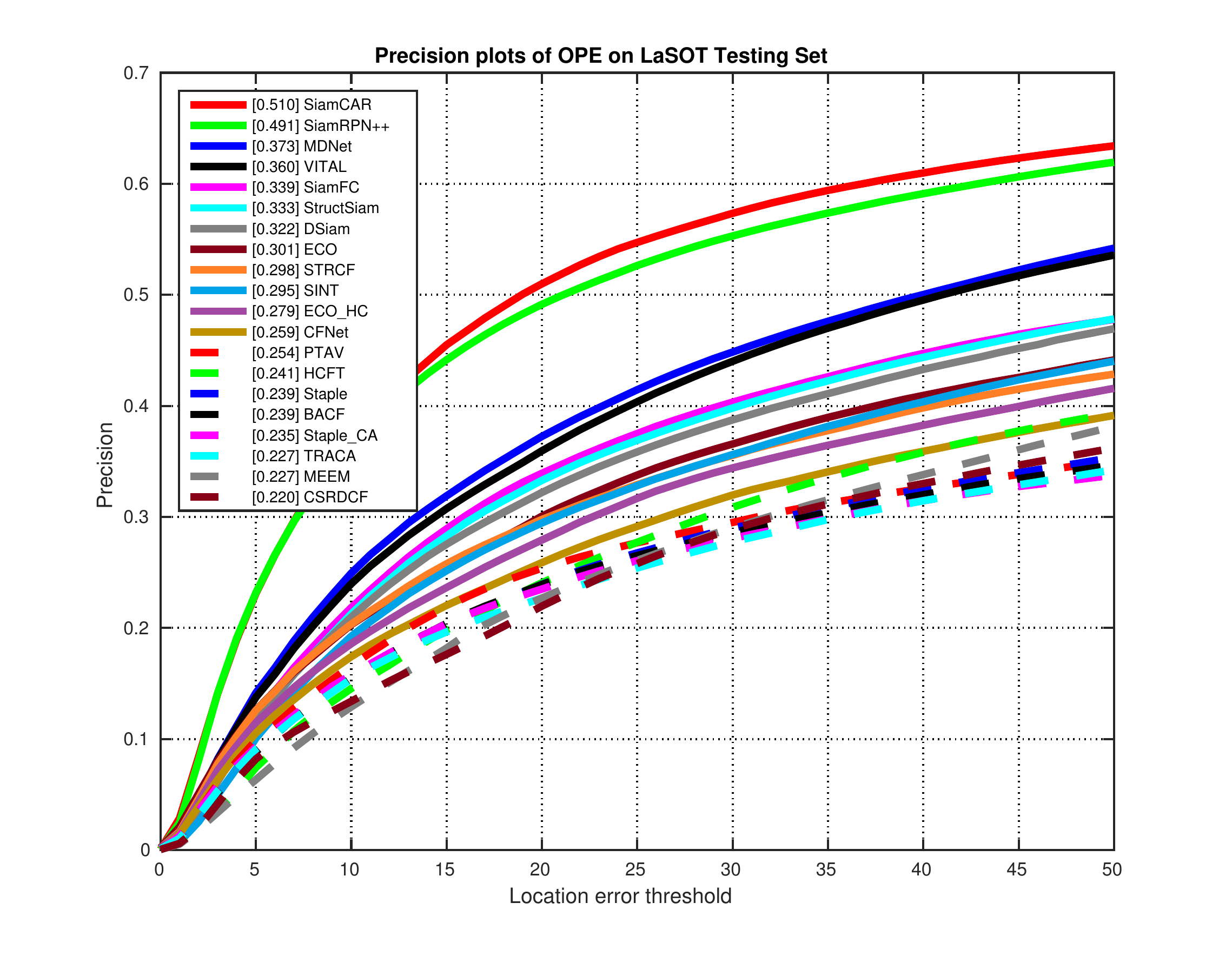}
		\end{minipage}
	}%
	\subfigure{
		\begin{minipage}[t]{0.31\linewidth}
			\centering
			\includegraphics[width=2.1in]{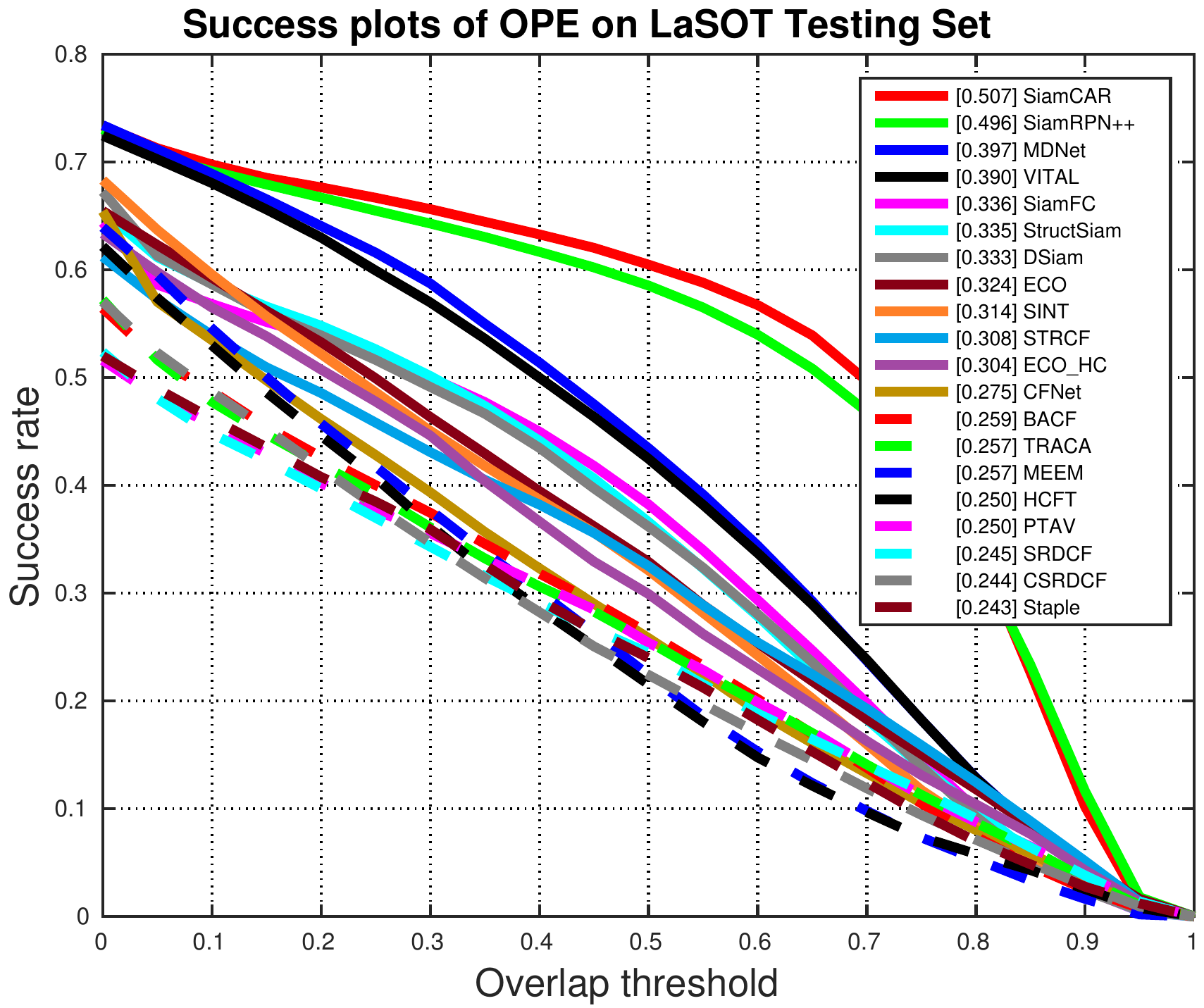}
		\end{minipage}
	}%
	\centering
	\caption{ Comparisons among Top-20 trackers on LaSOT \cite{lasot}. Our SiamCAR significantly outperforms the baselines and state-of-the-art methods.}
	\label{lasot}
\end{figure*}

\textbf{Training details}. During the training process, the batch size is set as $96$ and totally $20$ epochs are performed by using stochastic gradient descent (SGD) with an initial learning rate $0.001$. For the first $10$ epochs, the parameters of the Siamese subnetwork are frozen while training the classification and regression subnetwork. For the last $10$ epochs, the last $3$ blocks of ResNet-50 are unfrozen to be trained together. The whole training phase takes around $42$ hours. We train our SiamCAR with the data from COCO \cite{lin2014microsoft}, ImageNet DET, ImageNet VID \cite{russakovsky2015imagenet} and YouTube-BB \cite{real2017youtube} for experiments on GOT-10K \cite{got10k} UAV, OTB \cite{otb} and LaSOT \cite{lasot}. It should be noticed that for experiments on GOT-10K and LaSOT, our SiamCAR is trained with only the specified training set provided by the official website for fair comparison.

\textbf{Testing details}. During the testing process, we take use of the offline tracking strategy. Only the object in the initial frame of a sequence is adopted as the template patch. Consequently, the target branch of the Siamese subnetwork can be pre-computed and fixed during the whole tracking period.
The search region in the current frame is adopted as the input of the search branch. In Figure \ref{trackflow} we show a whole tracking process. With the outputs of classification-regression subnetwork, a location $q$ is queried through Equation (\ref{getq}).
In order to achieve a more stable and smoother prediction between adjacent frames, a weighted average of regression boxes corresponding to the top-3 neighbors of $q$  is computed as the final tracking result.

\begin{figure*}[htbp]
	\centering
	\subfigure{
		\begin{minipage}[t]{0.232\linewidth}
			\centering
			\includegraphics[width=1.63in]{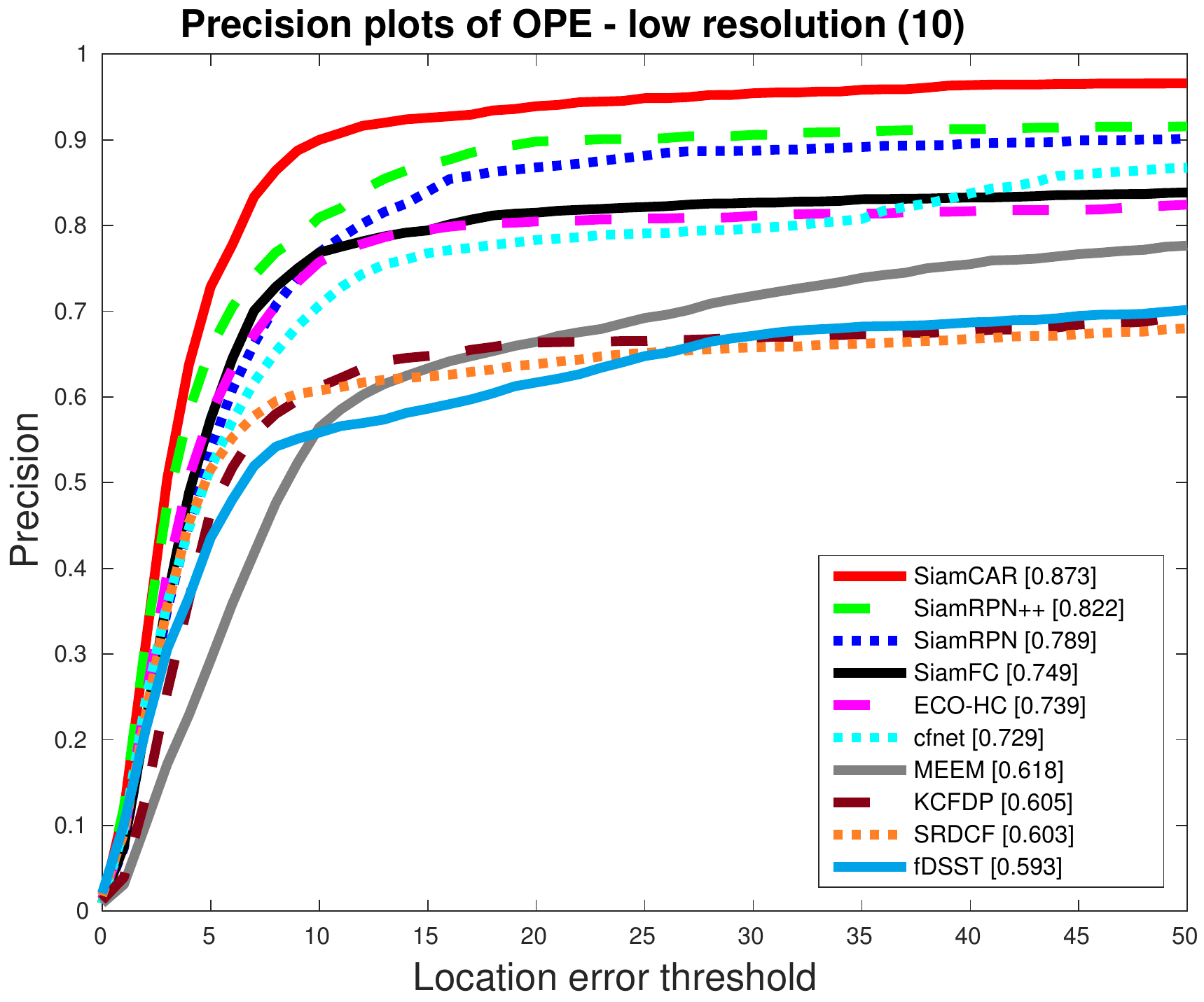}
		\end{minipage}%
	}%
	\subfigure{
		\begin{minipage}[t]{0.232\linewidth}
			\centering
			\includegraphics[width=1.63in]{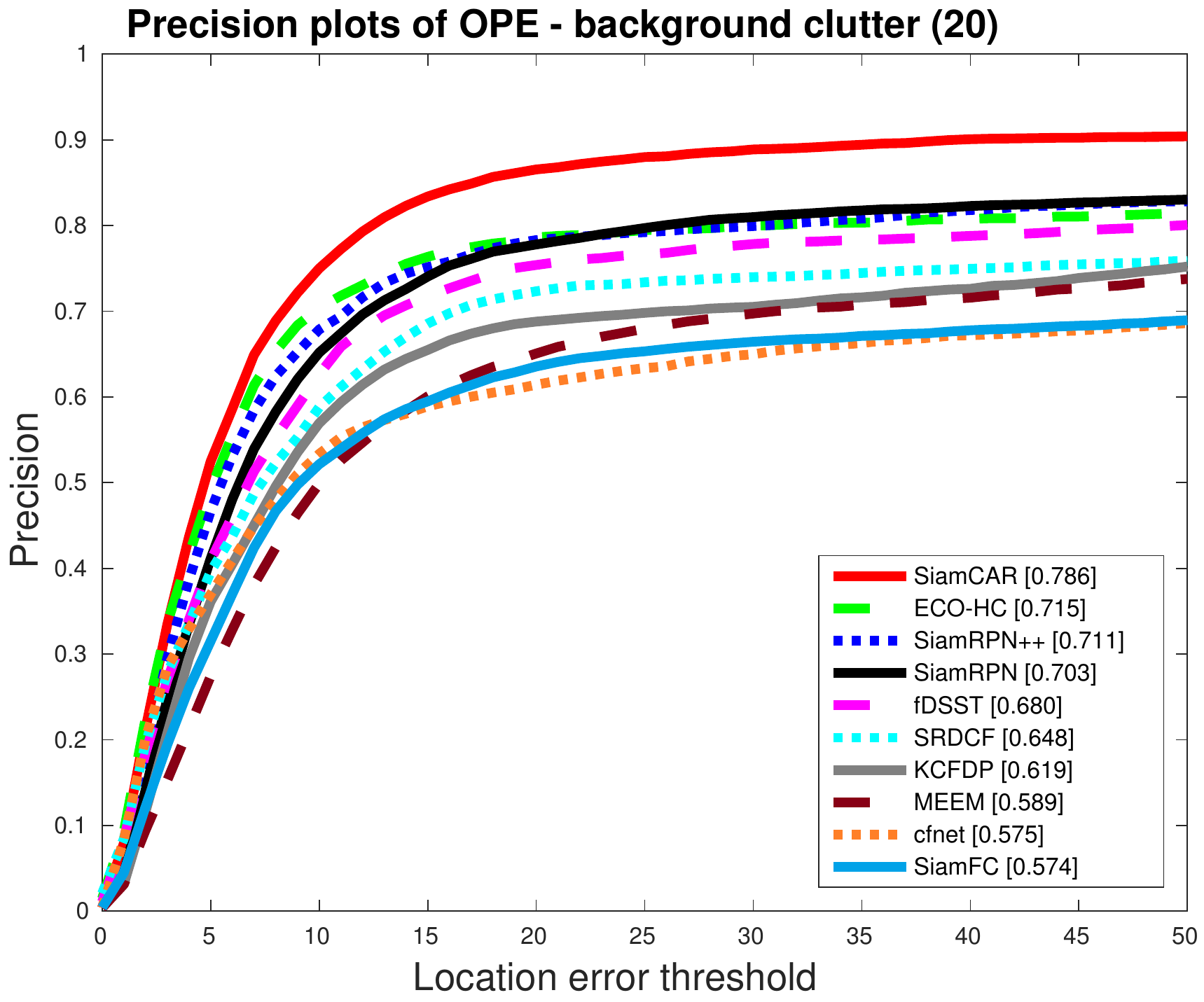}
		\end{minipage}
	}%
	\subfigure{
		\begin{minipage}[t]{0.232\linewidth}
			\centering
			\includegraphics[width=1.63in]{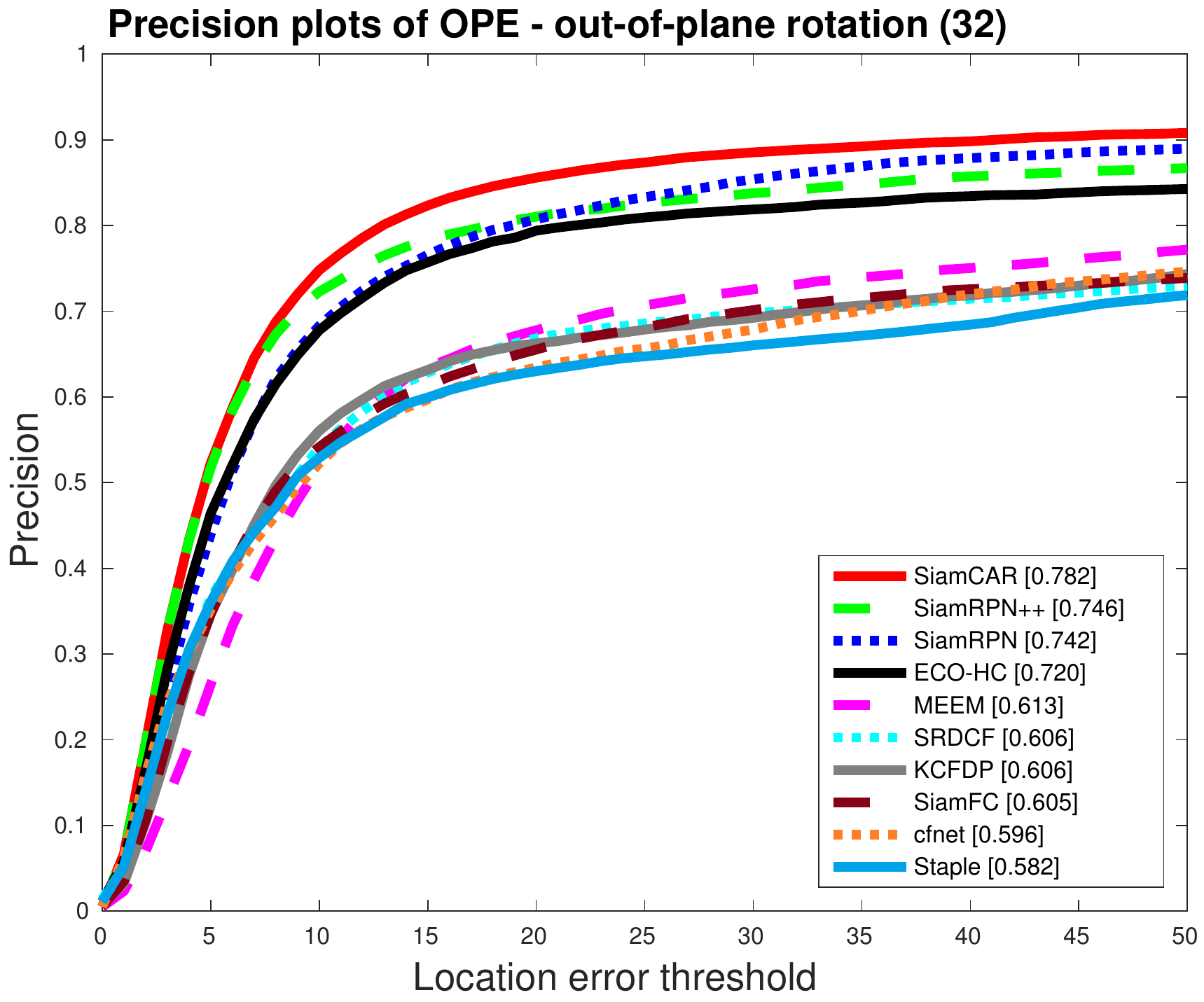}
		\end{minipage}
	}%
	\subfigure{
		\begin{minipage}[t]{0.232\linewidth}
			\centering
			\includegraphics[width=1.63in]{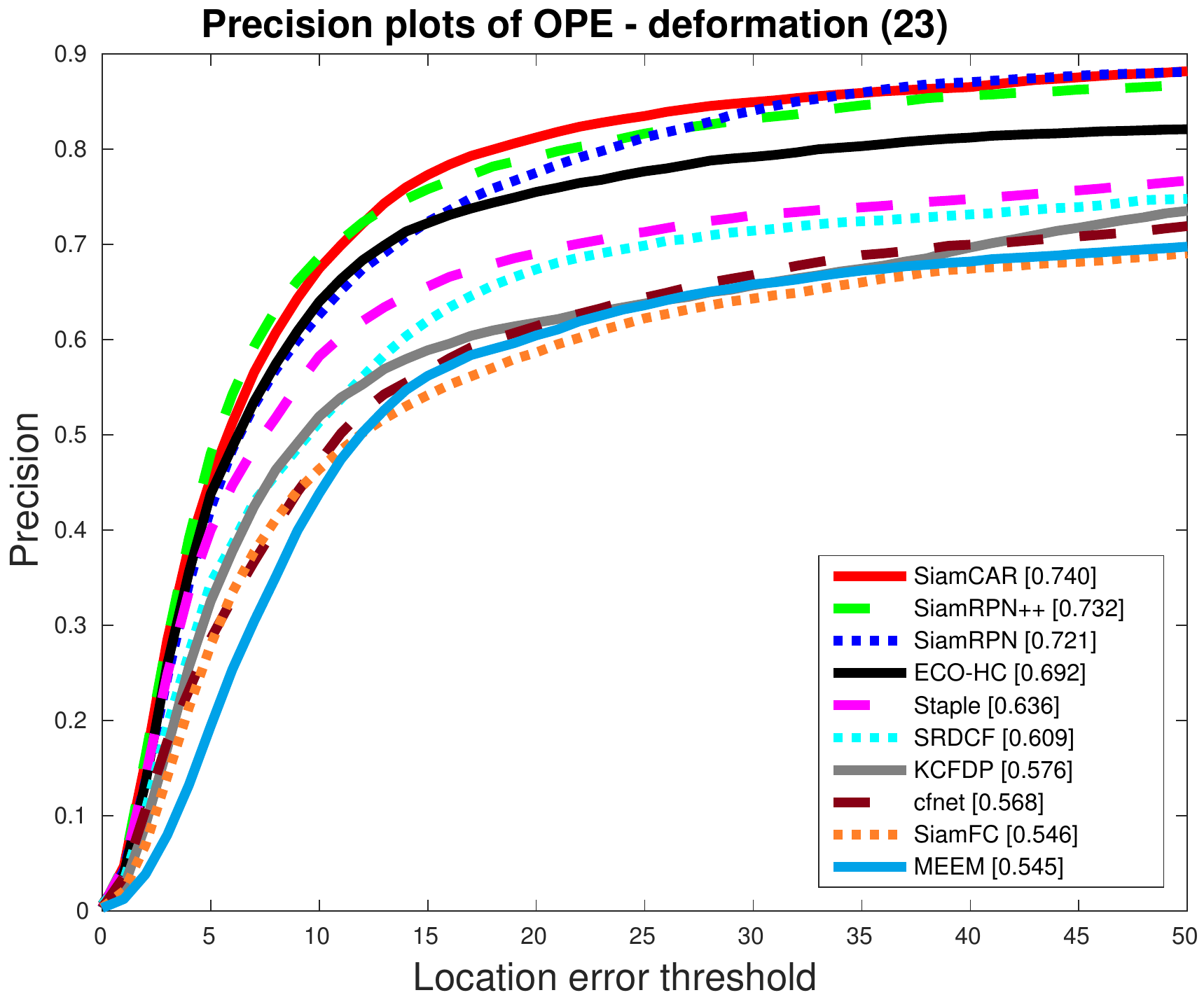}
		\end{minipage}
	}%
	
	\subfigure{
		\begin{minipage}[t]{0.232\linewidth}
			\centering
			\includegraphics[width=1.63in]{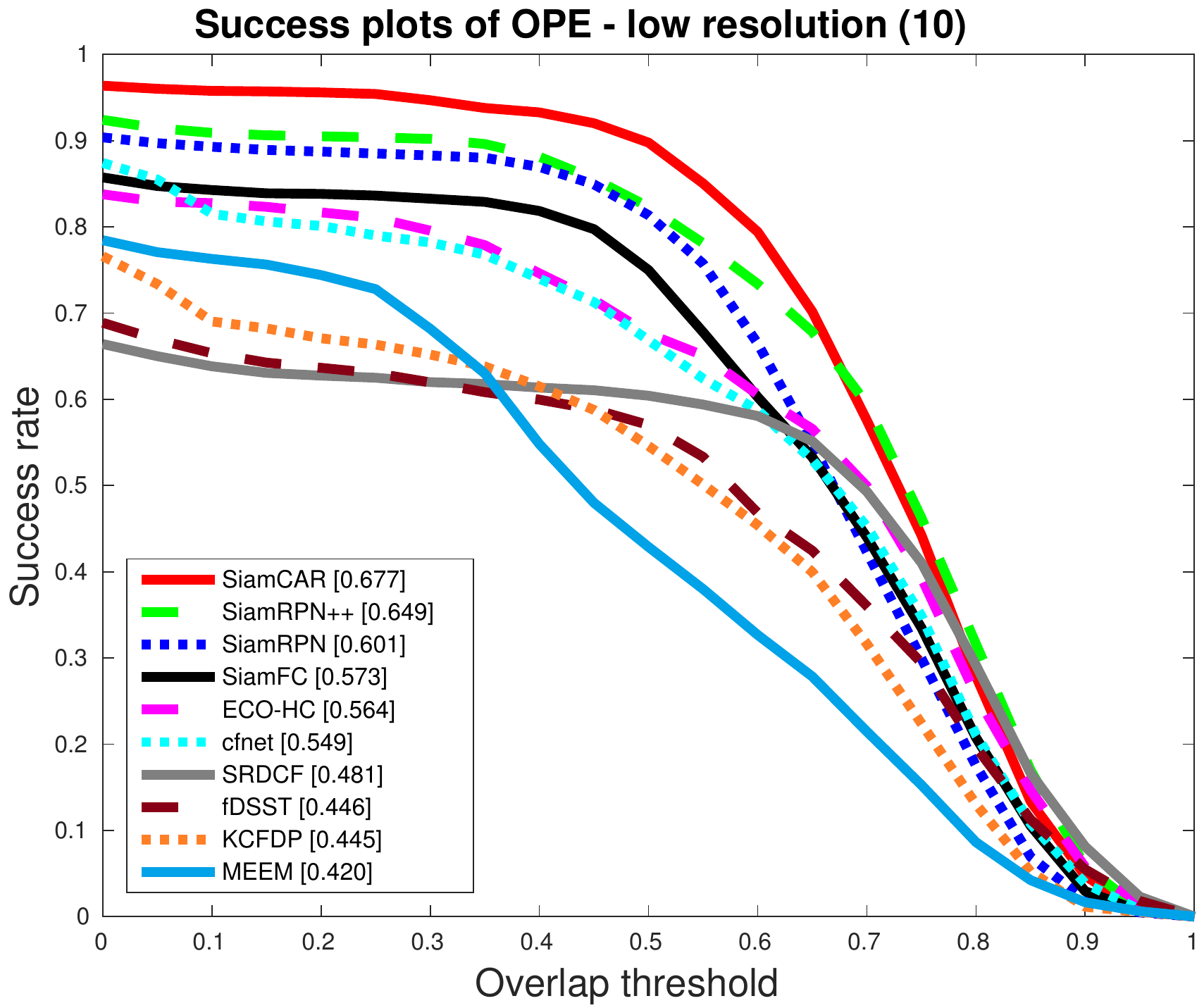}
		\end{minipage}%
	}%
	\subfigure{
		\begin{minipage}[t]{0.232\linewidth}
			\centering
			\includegraphics[width=1.63in]{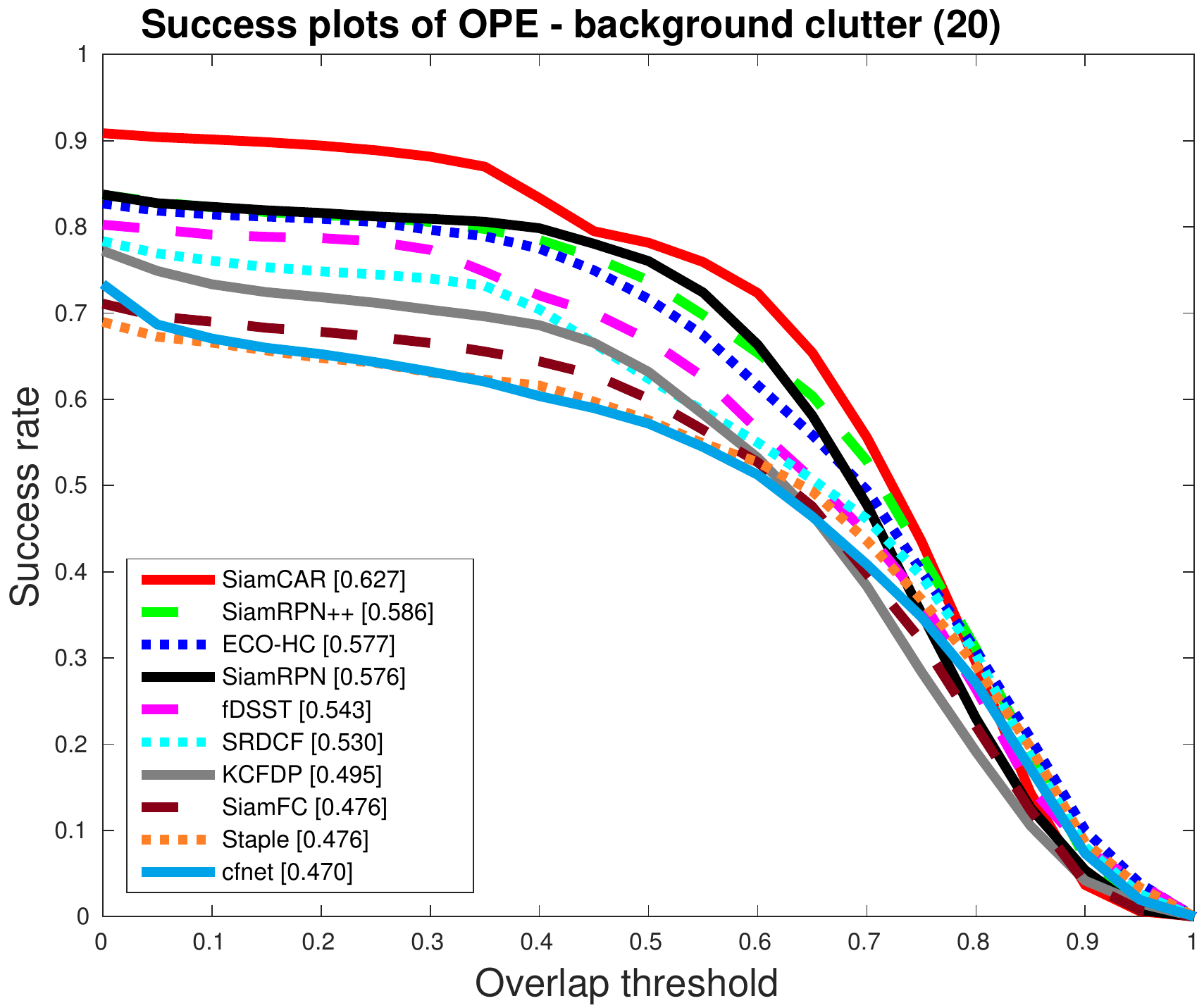}
		\end{minipage}
	}%
	\subfigure{
		\begin{minipage}[t]{0.232\linewidth}
			\centering
			\includegraphics[width=1.63in]{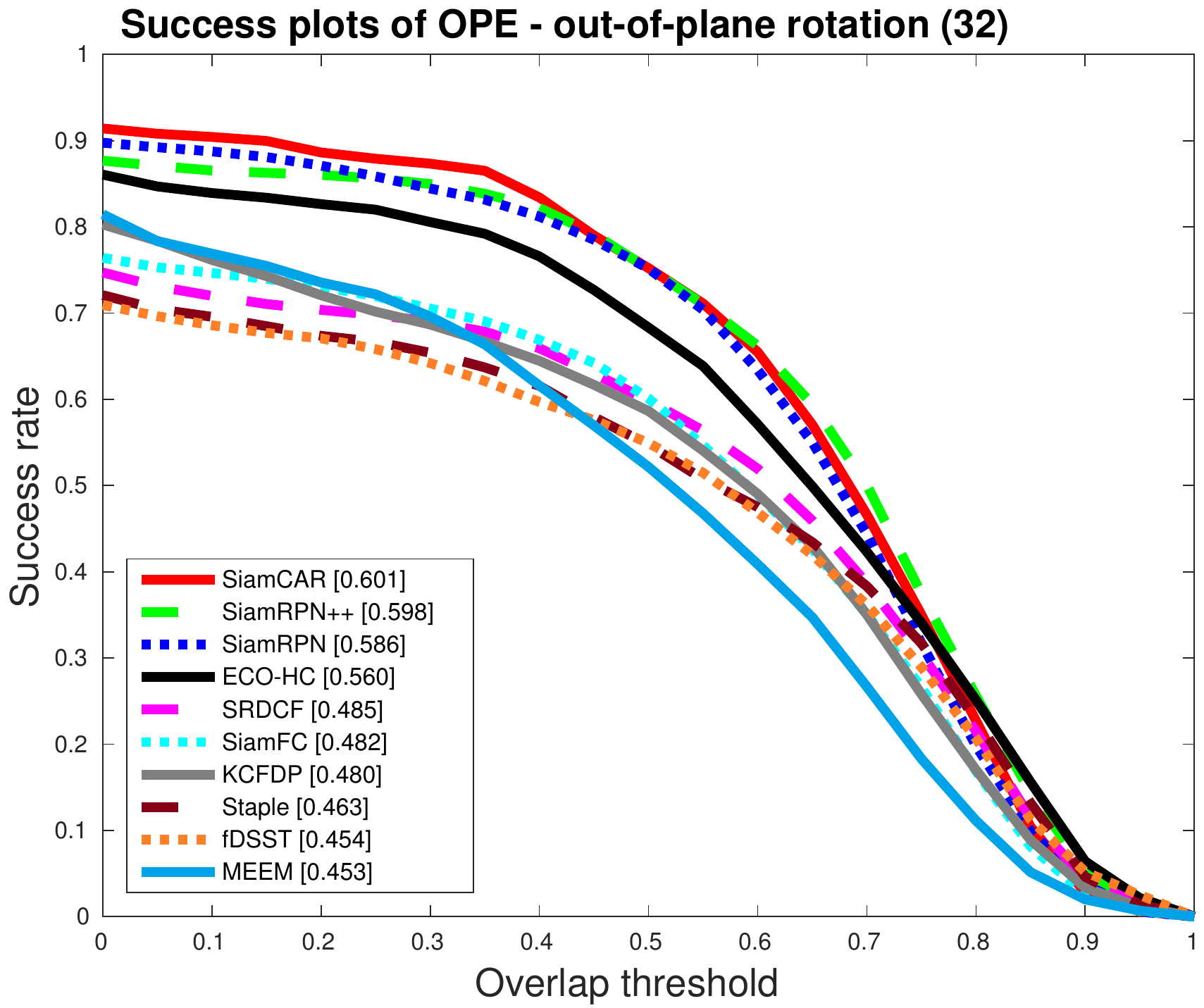}
		\end{minipage}
	}%
	\subfigure{
		\begin{minipage}[t]{0.232\linewidth}
			\centering
			\includegraphics[width=1.63in]{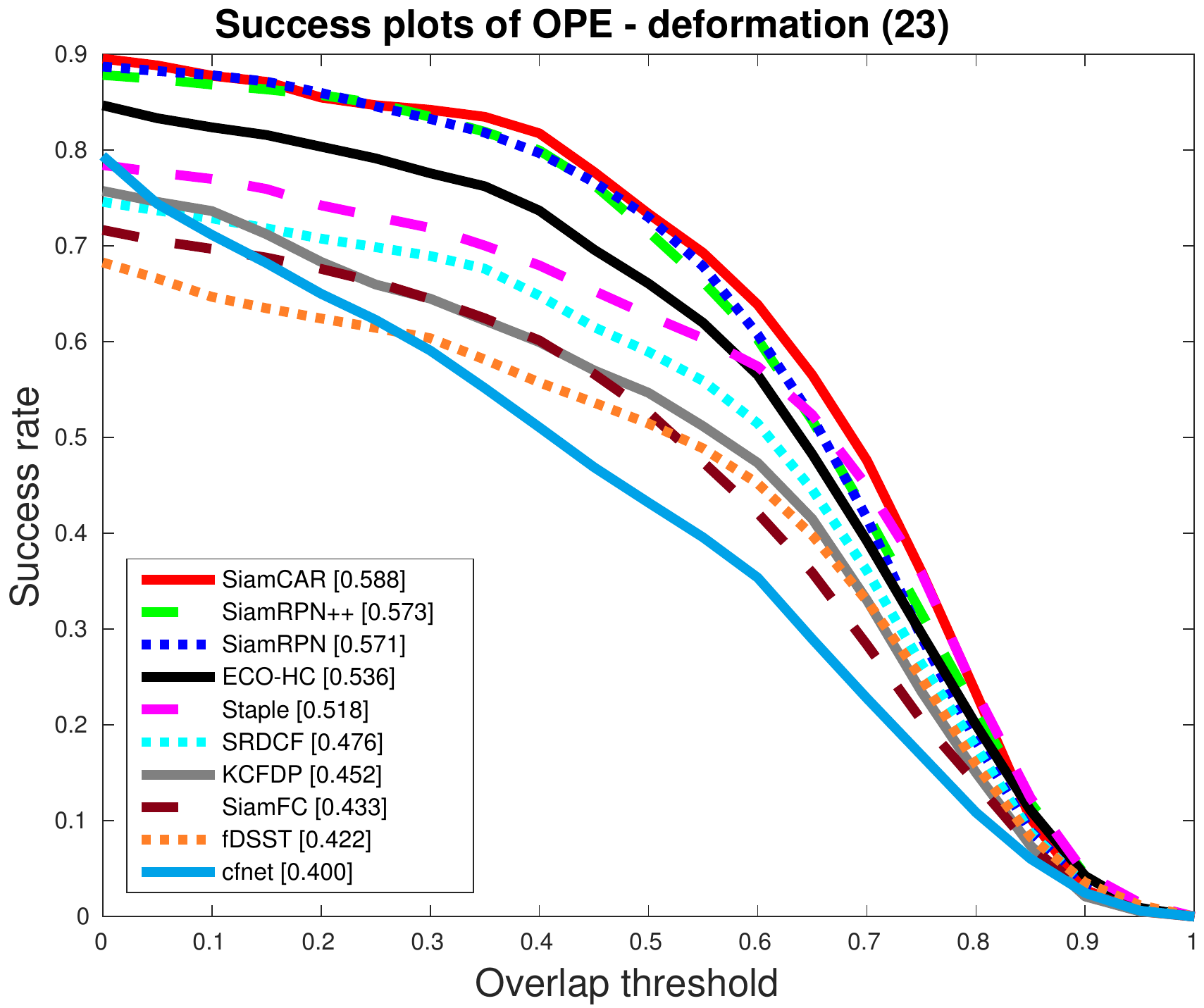}
		\end{minipage}
	}%
	\centering
	\caption{Comparisons on OTB-50 \cite{otb} with challenging aspects: low resolution, background clutter,  out-of-plane rotation and deformation. Our SiamCAR achieves the best results for all these aspects.}
	\label{otbcompare}
\end{figure*}

\subsection{Results on GOT-10K}
GOT-10K \cite{got10k} is a recently released large high-diversity benchmark for generic object tracking in the wild. It contains more than $10,000$ video segments of real-world moving objects. The fair comparison of deep trackers is ensured with the protocol that all approaches are using the same training data provided by the dataset. The classes in training dataset and testing dataset are zero overlapped. Authors need to train their models on the given training dataset and test them on the given testing dataset. After uploading the tracking results, the analysis is taken automatically by the official website. The provided evaluation indicators include success plots, average overlap ($AO$) and success rate ($SR$). The $AO$ represents the average overlaps between all the estimated bounding boxes and ground-truth boxes. The $SR_{0.5}$represents the rate of successfully tracked frames whose overlap exceeds $0.5$, while $SR_{0.75}$ represents the rate of successfully tracked frames whose overlap exceeds $0.75$.

 We evaluate SiamCAR on GOT-10K and compare it with state-of-the-art trackers including SiamRPN++ \cite{siamrpn++}, SiamRPN \cite{SiamRPN}, SiamFC \cite{SiamFC}, ECO \cite{ECO}, CFNET \cite{cfnet}  and other baselines or state-of-the art approaches.  All the results are provided by the official website of GOT-10K. Figure \ref{compGOT10K} shows that SiamCAR can outperforms all the trackers on GOT-10K and Table \ref{compGOT10K} lists the comparison details of different indicators. As shown in Table \ref{compGOT10K}, our tracker ranks 1st in terms of all the indicators. Compared with SiamRPN++, our SiamCAR improves the scores by $5.2\%$, $5.4\%$ and $9.0\%$ relatively for $AO$, $SR_{0.5}$  and $SR_{0.75}$. 

Since the trackers fairly use the same training data and the ground-truth boxes of the testing dataset are unseen for trackers, the tracking results on GOT-10K are more credible and convincing than those on other benchmarks.

\subsection{Results on LaSOT}
LaSOT is a resent released benchmark for single object tracking. The dataset contains more than $3.52$ million manually annotated frames and $1400$ videos. It contains $70$ classes and each class include $20$ tracking sequences. Such a large test dataset brings a great challenge to the tracking algorithms. The official website of LaSOT provides $35$ algorithms as baselines. Normalized precision plots, precision plots and success plots in one-pass evaluation ($OPE$) are considered as the indicators.

We compare our SiamCAR with the top-19 trackers including SiamRPN++ \cite{siamrpn++},MDNet \cite{mdnet}, DSiam \cite{DSiam}, ECO \cite{ECO} and other baselines. The results of SiamRPN++ \cite{siamrpn++} are provided on the website of its authors, while other results are provided by the official website of LaSOT. As shown in Figure \ref{lasot}, our SiamCAR achieves the best performance. Compared with SiamRPN++, our SiamCAR improve the scores by $3.1\%$, $1.9\%$ and $1.1\%$ relatively for the three indicators. Notably, compared with the provided baselines, our SiamCAR make a great progress by improving the scores by over  $14\%$, $13.7\%$ and $11\%$ relatively for the three indicators.

The leading results on such a large dataset demonstrate that our proposed network has a good generalization for visual object.


\begin{figure*}[htbp]
	\centering
	\subfigure{
		\begin{minipage}[t]{0.50\linewidth}
			\centering
			\includegraphics[width=3.1in]{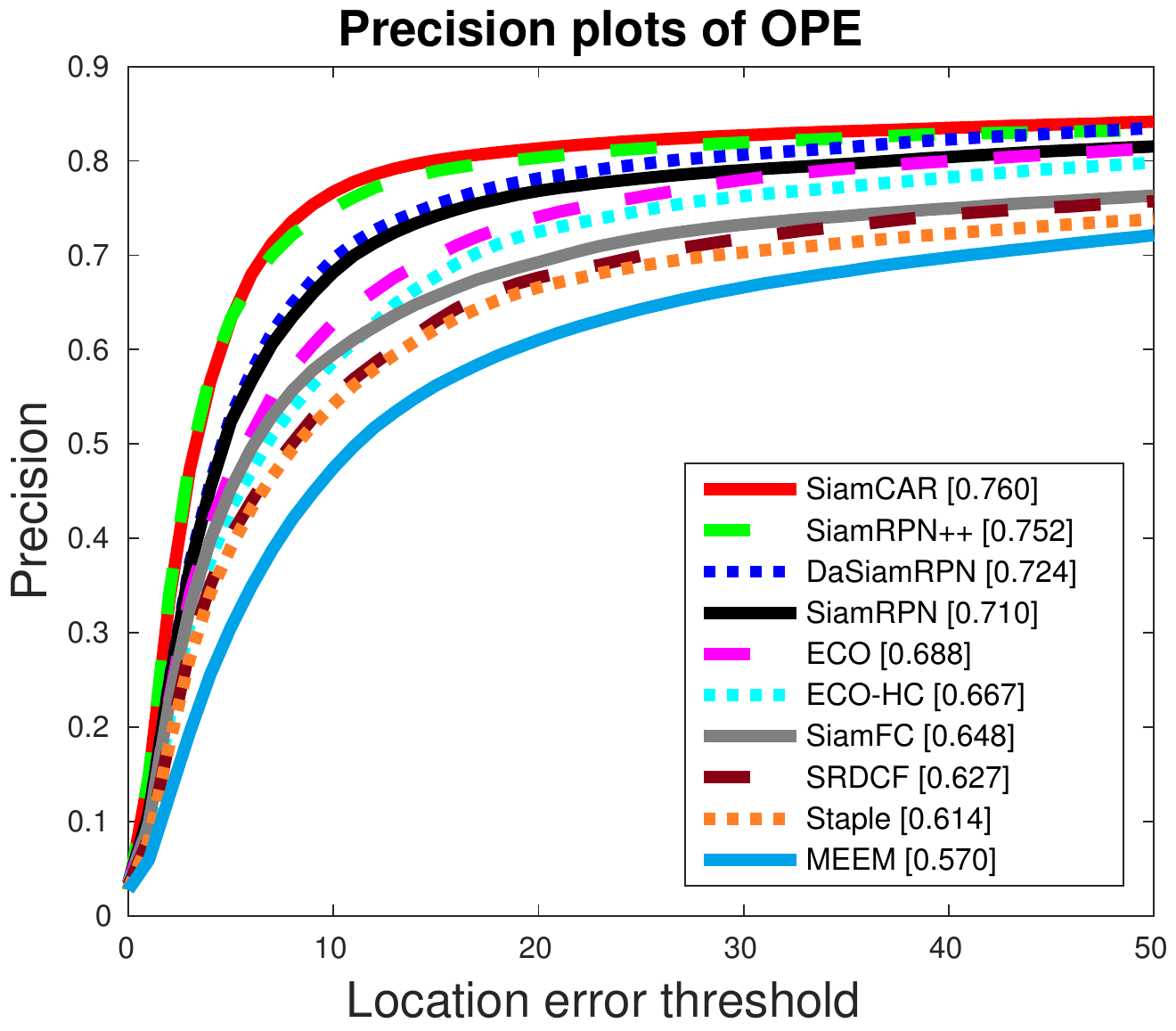}
		\end{minipage}%
	}%
	\subfigure{
		\begin{minipage}[t]{0.50\linewidth}
			\centering
			\includegraphics[width=3.2in]{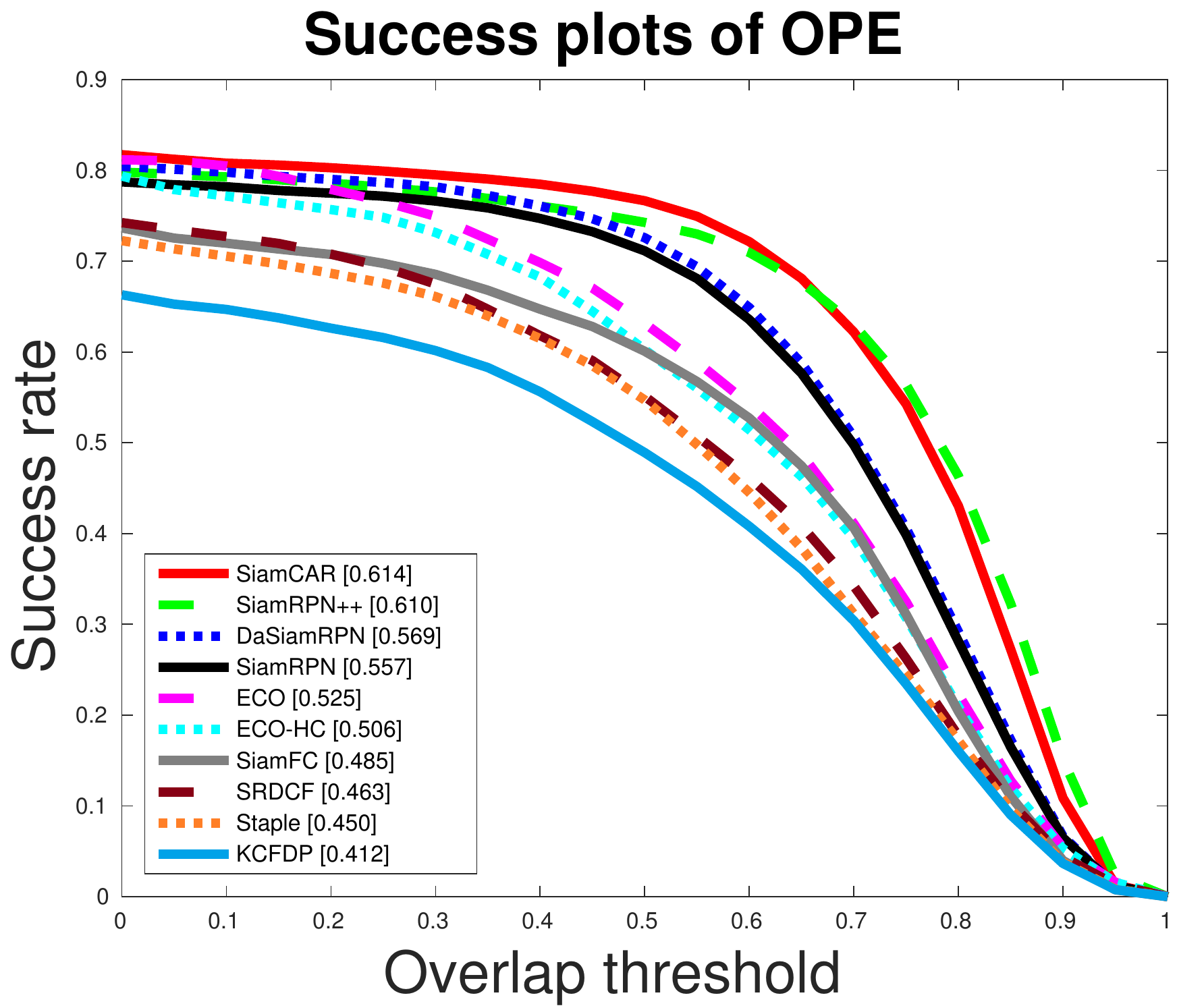}
		\end{minipage}
	}%
	\centering
	\caption{Comparisons on UAV123. Our SiamCAR achieves the best results.}
	\label{uav123}
\end{figure*}

\subsection{Results on OTB50}
OTB-50 contains $50$ challenging videos with substantial variations. The test sequences are manually tagged with $9$ attributes to represent the challenging aspects, including illumination variation, scale variation, occlusion, deformation, motion blur, fast motion, in-plane rotation, out-of-plane rotation, out-of-view, background clutters and low resolution. We compare our network with 9 state-of-the-art approaches including SiamRPN++ \cite{siamrpn++},  SiamRPN \cite{SiamRPN}, SiamFC \cite{SiamFC} and ECO \cite{ECO}. The success plots and precision plots in $OPE$ for each tracker are evaluated. As shown in Figure \ref*{otbcompare}, the proposed SiamCAR ranks 1st in terms of both two indicators with those challenging aspects.
Especially, our SiamCAR significantly improves the tracking accuracy for the aspects of low resolution, out-of-plane rotation and background clutter. The results demonstrate that SiamCAR can better deal with the challenging distractors and large pose variation, which benefit from the implicitly decoded semantic information of our classification-regression subnetwork.

\subsection{Results on UAV123}



UAV123 dataset contains in total of $123$ video sequences, including more than $110K$ frames. All sequences are fully annotated with upright bounding boxes. The objects in the dataset mainly suffer from fast motion, large scale variation, large illumination variation and occlusions, which make the tracking challenging. 

We compare our SiamCAR with 9 state-of-the-art approaches including SiamRPN++ \cite{siamrpn++},  SiamRPN \cite{SiamRPN}, SiamFC \cite{SiamFC} and ECO \cite{ECO} on this dataset. The success plot and precision plot of OPE are used as indicators to evaluate the overall performance. As shown in Figure  \ref{uav123}, our SiamCAR outperforms all other trackers for both indicators. Compared with state-of-the-art RPN trackers \cite{siamrpn++, DaSiamRPN, SiamRPN}, SiamCAR obtains competitive results with much simple network and without heuristic tuning parameters.

\subsection{Run-time Evaluation}
In column FPS of Table \ref{compGOT10K}, we show the evaluation on GOT-10K in respect to the frame-per-second (FPS). The reported speed is evaluated on a machine with one RTX 2080ti and others are provided by the GOT-10K official results. As shown in the table, our SiamCAR achieves the best performance at a real-time speed with $52.27$ FPS. In addition, our network is much simpler than others and no special designed parameters are needed for training.

\section{Conclusions}
In this paper, we present a Siamese classification and regression framework as called SiamCAR to end-to-end train a deep Siamese network for visual tracking. We show that the tracking task can be solved in a per-pixel manner and adopted with the neat fully convolution framework. The proposed framework is very simple in structure but achieves state-of-the-art results without bells and whistles on GOT-10K and many other challenging benchmarks. It also achieves state-of-the-art results on large dataset such as LaSOT, which demonstrate the generalizability of our SiamCAR. Since the present framework is simple and neat, it can be easily to be modified with specific modules to make further improvement in the future.


{\small
\bibliographystyle{ieee_fullname}
\bibliography{egpaper_for_review}
}

\end{document}